\documentclass{article}
\usepackage{spconf,amsmath,graphicx}
\usepackage{graphicx}
\usepackage{amsmath} 
\usepackage{kotex}
\usepackage{xcolor}
\usepackage{multirow}

\title{A novel approach for holographic 3D content generation without depth map}
\begin{document}
%
\maketitle
\begin{abstract}
In preparation for observing holographic 3D content, acquiring a set of RGB color and depth map images per scene is necessary to generate computer-generated holograms (CGHs) when using the fast Fourier transform (FFT) algorithm. However, in real-world situations, these paired formats of RGB color and depth map images are not always fully available. We propose a deep learning-based method to synthesize the volumetric digital holograms using only the given RGB image, so that we can overcome environments where RGB color and depth map images are partially provided. The proposed method uses only the input of RGB image to estimate its depth map and then generate its CGH sequentially. Through experiments, we demonstrate that the volumetric hologram generated through our proposed model is more accurate than that of competitive models, under the situation that only RGB color data can be provided.
\end{abstract}
\section{Introduction}
\label{sec:intro}

Computer-Generated Holograms (CGH) are generated from RGB images coupled with their corresponding depth maps. While such input data can be acquired by specific camera products, there exists pairwise inconsistency in their resolution. This leads to CGH generation methods requiring a preprocessing step that involves aligning the resolution of a given RGB image and depth map. Moreover, providing realistic 3D holographic content demands high resolution (2K ($1920\times1080$) and 4K ($3840\times2160$)) RGB image-depth map pair. Such processes incur large computational costs. To relieve such a burden, we propose an approach that circumvents this preprocessing step by generating holographic 3D content only from RGB images.

\begin{figure}[th!]

\begin{minipage}[t]{1.0\linewidth}
  \centering
  \centerline{\includegraphics[width=0.92\columnwidth]{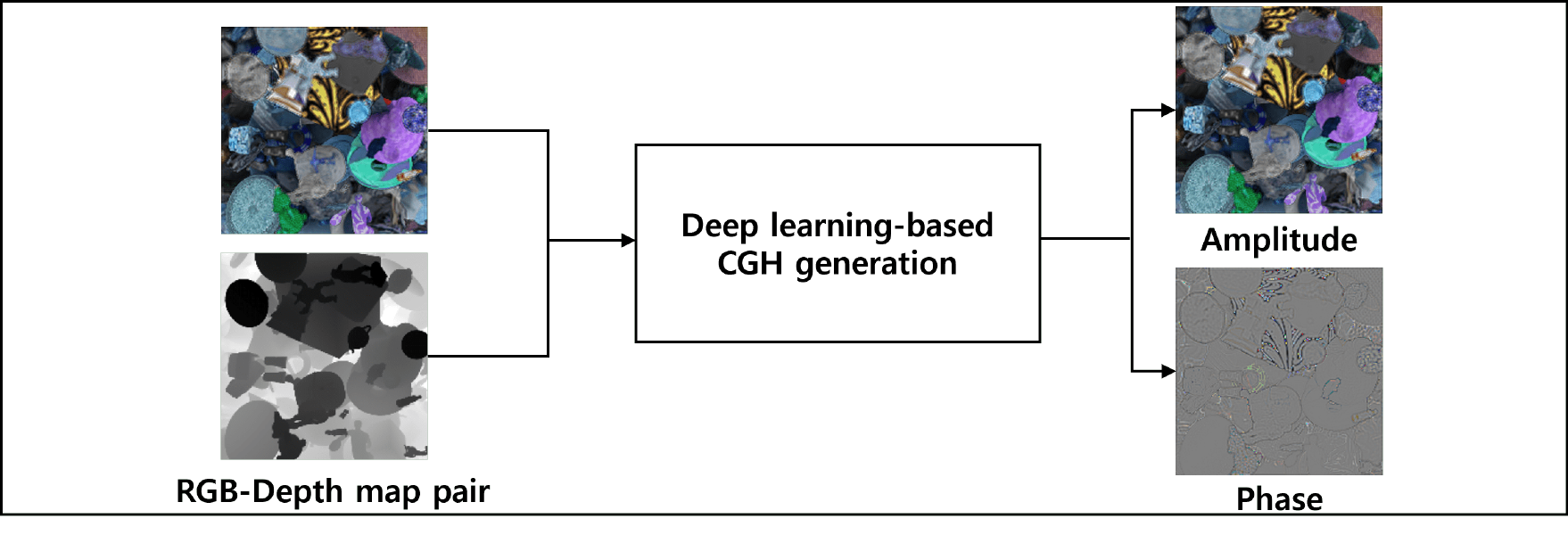}}
  \centerline{(a)}\medskip
\end{minipage}
\begin{minipage}[t]{1.0\linewidth}
  \centering
  \centerline{\includegraphics[width=0.92\columnwidth]{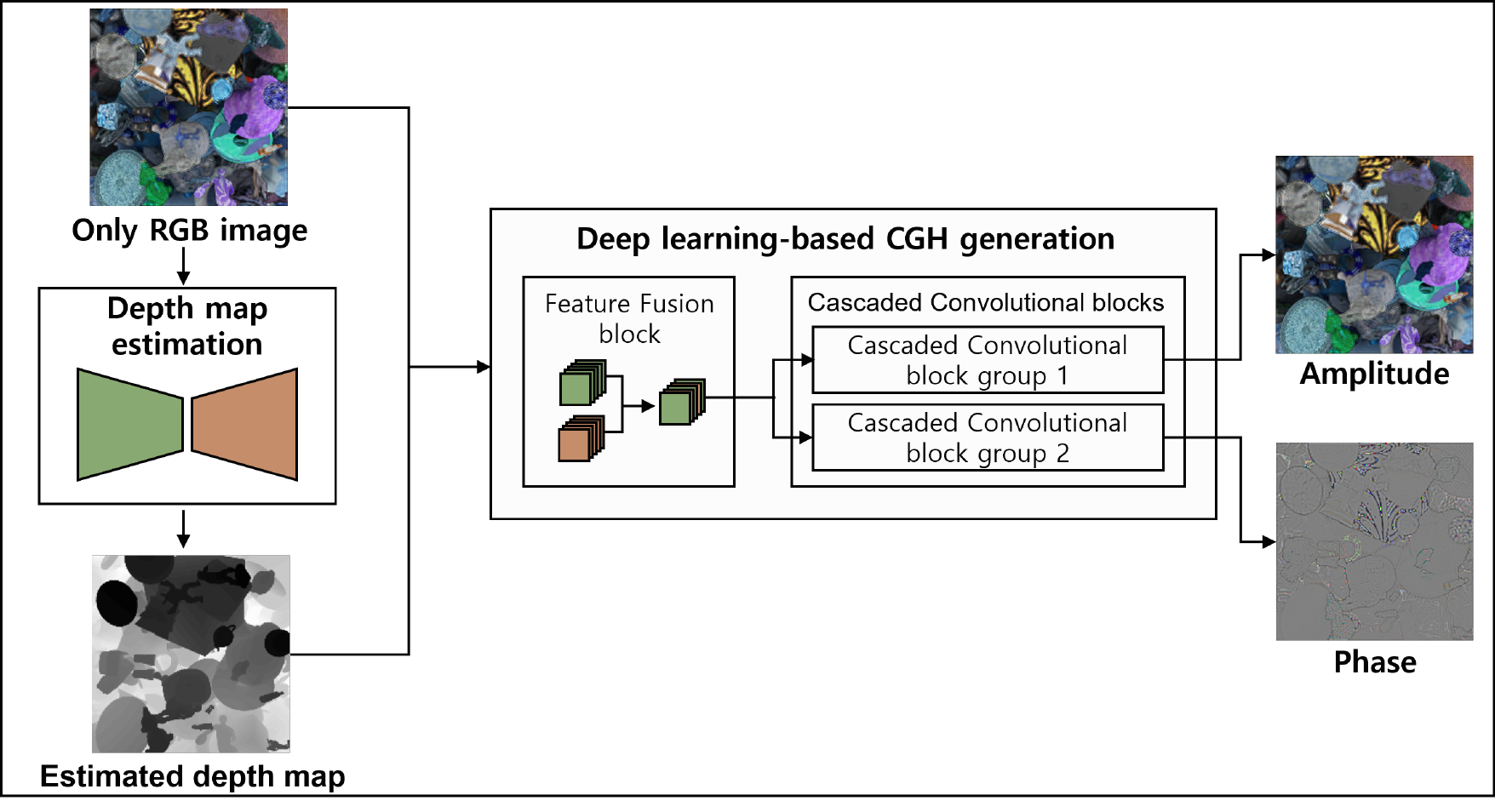}}
  \centerline{(b)}\medskip
\end{minipage}

\caption{Overview of the proposed method. Conventional CGH generation methods (a): Both RGB image and depth map are required to generate holographic 3D content. Proposed method (b): Only RGB image is required to generate holographic 3D content once depth map learning is complete.}
\label{fig1}
\end{figure}

Figure~\ref{fig1} shows the differences between the conventional CGH generation methods (a) and our newly proposed method (b). Our proposed method consists of an Embedded Depth map Estimation module and a CGH Generation module. The Embedded Depth map Estimation module first estimates a depth map using an RGB image while the CGH Generation module subsequently generates the volumetric CGH. Experimental results show that the quality of volumetric CGHs generated by our approach does not fall behind those of other state-of-the-art models, which use both types of input. 

\section{Related work}
\subsection{Assistance module in computer vision}
While our Embedded Depth map Estimation module estimates a depth map used for generating a volumetric 3D hologram, previous works have employed similar approaches by using assistance modules. Zhang et al. used a guidance module to predict segmentation images~\cite{zhang2019net}. Nazeri et al. used assistance modules in image inpainting to generate a complete image~\cite{nazeri2019edgeconnect} with a Generative Adversarial Network~\cite{goodfellow2014generative}-based model. Huang et al. and Jiao et al. proposed methods that intermediately predict depth maps to achieve their purpose\cite{huang2019wireframe, jiao2019geometry}. Zhang et al. used a U-Net-based~\cite{ronneberger2015u} model that jointly output segmentation and depth map estimation~\cite{zhang2018joint}. Wang et al. augmented depth map prediction process with a semantic segmentation module~\cite{wang2020sdc}. Kumar et al. proposed a framework for classifying dynamic objects in driving situations which features acquiring the segmentation model and utilizing them as guidance features~\cite{kumar2021syndistnet}.

\subsection{Digital Hologram Generation}
Studies related to digital hologram generation can be categorized into generating holograms for 2D and 3D scenes. Long at el. utilized FCNs (Fully Convolution Networks) and GAN to generate 2D holograms~\cite{long2015fully}. Khan et al. proposed a GAN-based model that generates CGHs quickly~\cite{khan2021gan}. Lee et al. presented the method to generate holograms using distance information~\cite{lee2020deep}. Shi et al. introduced an approach for generating volumetric 3D holograms using an RGB image and depth image~\cite{shi2021towards}. The novelty of our approach is based on utilizing only RGB images to generate volumetric 3D digital holograms which is a distinguishable feature from other previous works.

\section{Proposed method}
\subsection{Model Architecture}
The proposed model architecture consists of two modules which are the Embedded Depth map Estimation module and the CGH Generation module. The Embedded Depth map Estimation module generates a depth map using an RGB image as input. The intermediate latent feature maps are propagated to the CGH Generation module. The CGH Generation module first fuses each layer-wise latent feature maps in Feature Fusion block and uses them to generate amplitude and phase through Cascaded Convolutional blocks.

\begin{figure*}[th!]
  \centering
   \includegraphics[width=\textwidth]{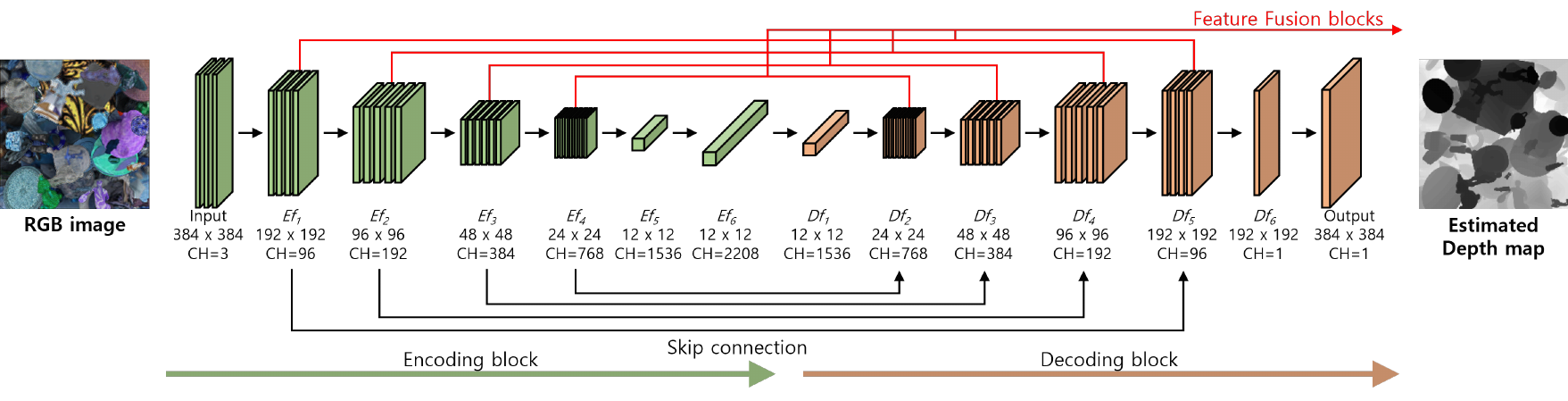}
   \caption{Embedded Depth map Estimation module. The Encoding block performs feature extraction and down-sampling, and the Decoding block performs up-sampling, and generates a depth map.}
   \label{fig2}
\end{figure*}

Figure~\ref{fig2} illustrates the Embedded Depth Map Estimation module. For the Encoding block which performs feature extraction and down-sampling, we imported DenseNet161~\cite{huang2017densely}. Given a $384\times384$ RGB image with 3 channels as input, the Encoding block repeatedly propagates it through 6 convolution filters to create a $192\times192$ feature map. The number of channels for each feature map is 96, 192, 384, 768, 1536, and 2208 respectively while all convolution filters' size and stride are identically set to $3\times3$ and 1 respectively.

The Decoding block performs up-sampling and estimates a depth map using the lastly extracted feature map as input. The Embedded Depth map Estimation module uses skip connections, which reduces the incurring spatial information loss in each successive layer. We connected 4 encoding layers' feature maps to 4 decoding layers' feature maps. The Encoding block-Decoding block's latent feature maps ($Ef_n$, $Df_n$, $1 \leq n \leq 4$ in Figure~\ref{fig2}) are used as inputs for the Feature Fusion blocks. The Embedded Depth Map Estimation module is defined as

\begin{equation}
  Ef_{n+1} = \sum_{x=0}^{h-1}\sum_{y=0}^{w-1}Ef_{n}(x,y)*k
  \label{encoder}
\end{equation}

\begin{equation}
  Df_{n+1} = Bilinear(Up(Skip(Ef_{n},Df_{n})))
  \label{decoder}
\end{equation}

where $Ef$ is Encoding block's feature map, $h, w$ are size of width and height, $k$ is filter, and $Df$ is Decoding block's feature map. We used bilinear interpolation method in consideration of both computational cost and performance.

\begin{figure}[ht]
  \centering
   \includegraphics[width=0.6\columnwidth]{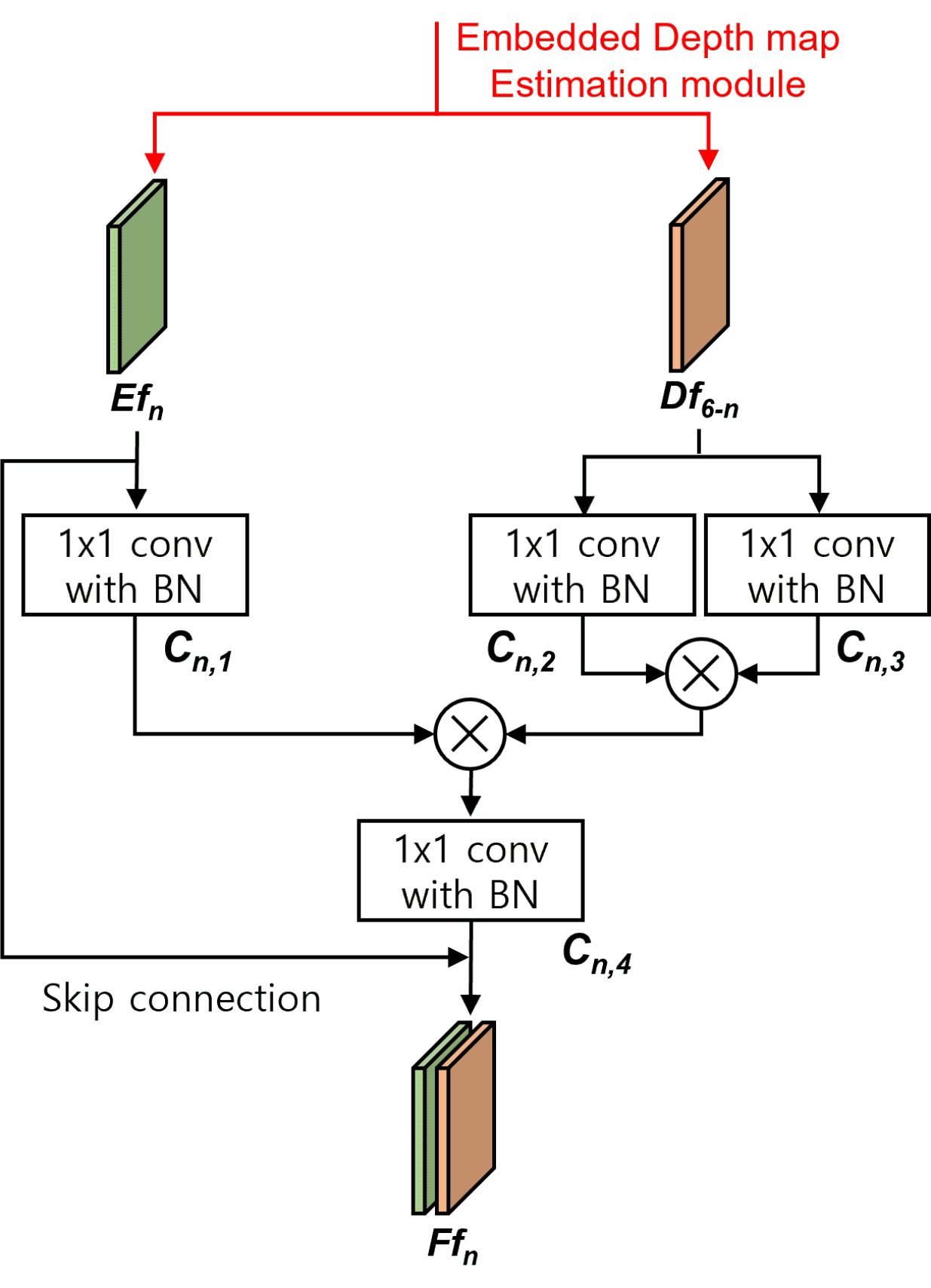}
   \caption{Feature Fusion block. Using the encoder-decoder's latent feature maps ($Ef_n$, $Df_n$, $1 \leq n \leq 4$) as inputs, a fused feature ($Ff_n$) is generated through $1\times1$ convolution, batch normalization (BN), element-wise product, and skip connection.}
   \label{fig3}
\end{figure}

The CGH Generation module consists of the Feature Fusion block and Cascaded Convolutional blocks. Figure~\ref{fig3} illustrates the Feature Fusion block. We imported Geometry-Aware Propagation (GAP) module~\cite{jiao2019geometry} as the Feature Fusion block. The GAP module generates features using a value at the same pixel location in two images to fuse two different pieces of information. The Feature Fusion block is defined as

\begin{equation}
\begin{split}
  Ff_{n} = Skip(Ef_{n}, C_{n,4}(C_{n,1}(Ef_{n})\times\\
  (C_{n,2}(Df_{n})\times C_{n,3}(Df_{n}))))
  \label{Ffn}
\end{split}
\end{equation}

\begin{equation}
  C_{n,m}(j) = BN(\sum_{x=0}^{h-1}\sum_{y=0}^{w-1}j(x,y)*(1,1))
  \label{conv}
\end{equation}

where $C$ is convolution operation with batch normalization (BN), $n, m$ are feature maps' step number and $C$s' number ($1 \leq n, m \leq 4$), $j$ is input feature map, $(1, 1)$ is 1 by 1 convolution operation. The multiplication symbol is an element-wise product and the asterisk is a vector product. The Feature Fusion block creates new features for generating a hologram. Fused feature maps ($Ff_n$ in Figure~\ref{fig3}) are used as input of the Cascaded Convolutional blocks.

\begin{figure}[htb]
  \centering
   \includegraphics[width=\columnwidth]{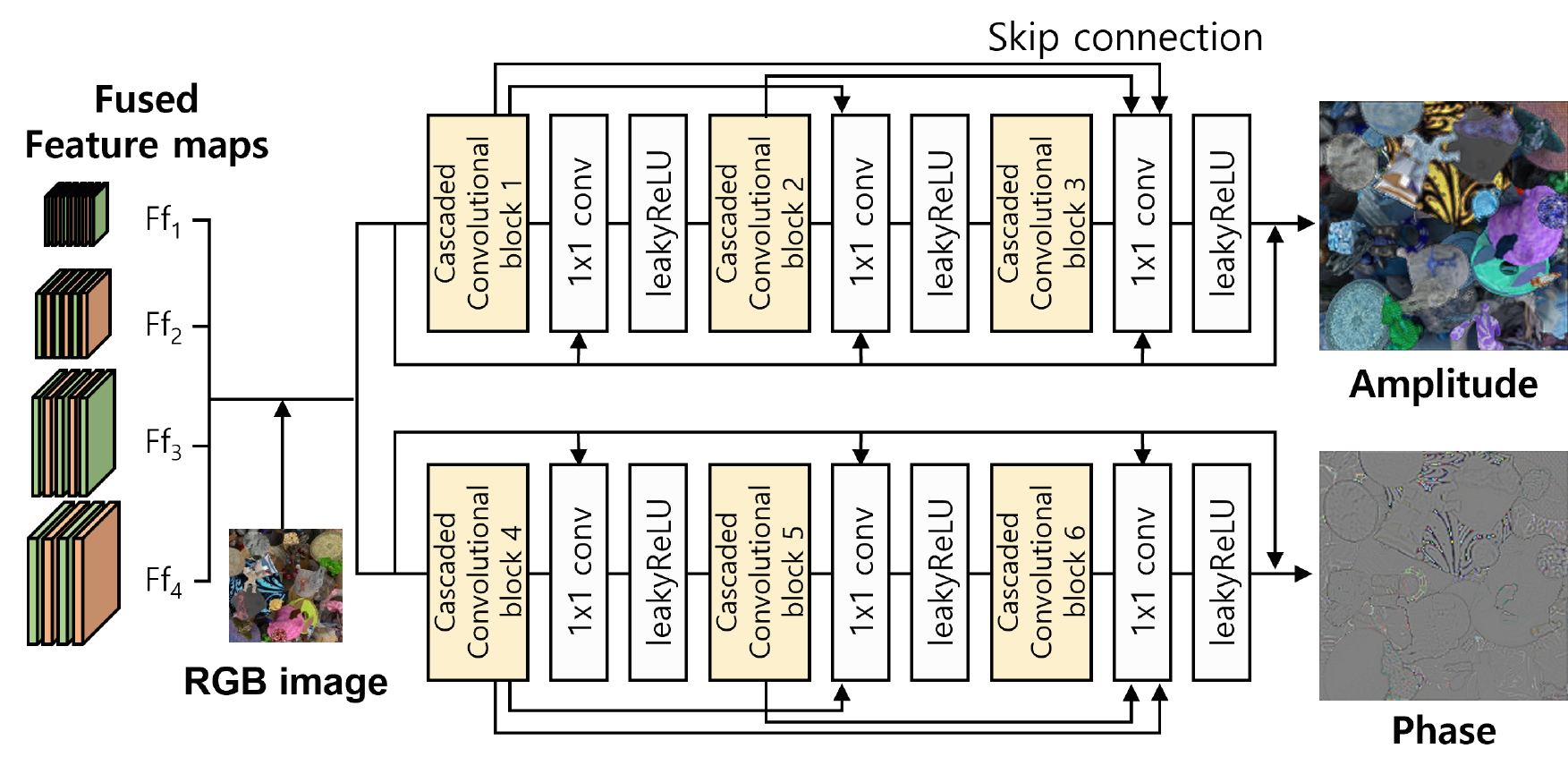}
   \caption{Cascaded Convolutional blocks. Using fused feature maps and an RGB image as input, 6 Cascaded Convolutional blocks, $1\times1$ convolution, and activation function are repeatedly used to generate amplitude and phase.}
   \label{fig4}
\end{figure}

Figure~\ref{fig4} illustrates the Cascaded Convolutional blocks. The fused feature maps and RGB images are used as input. 3 Cascaded Convolutional blocks are used to generate the amplitude while the 3 others are used to generate the phase. Each Cascaded Convolutional block consists of a $3\times3$ convolution layer, batch normalization, and an up-sampling layer, followed by a $1\times1$ convolution layer and a non-linear activation function LeakyReLU~\cite{xu2015empirical}.

\subsection{Model Optimization}
The optimization strategy for the proposed model consists of two phases which are depth map estimation and volumetric CGH generation respectively. In phase 1, the Embedded Depth Map Estimation module is trained under a loss criterion comprising MSE (Mean Squared Error) and SSIM (Structural Similarity) that minimizes the discrepancy between estimated and ground truth depth maps. In phase 2, the CGH Generation module is trained under a loss criterion comprising MSE and L1 Norm that minimizes the discrepancy between generated and ground truth amplitude-phase pairs. To determine loss criteria, we experimented by changing the coefficients $a1$ for SSIM and the coefficients $a2$ for L1 Norm, as a result, we found that the optimal values of $a1$ and $a2$ were 0.01.

\section{Experiments}
We performed two experiments to evaluate our proposed approach. The first experiment is constrained to generating volumetric holograms using only RGB images. Since other competitive models used in the experiment\cite{shi2021towards, lee2020deep, khan2021gan} require RGB-Depth pair, we fed them RGB images coupled with depth maps with pixels filled with zero depth values whereas we fed the proposed model only RGB images. The second experiment eliminates this constraint, granting other models\cite{shi2021towards, lee2020deep, khan2021gan} used in the experiment to fully use both sides of the input. 

\subsection{Dataset and Hyperparameter Setting}
We used the dataset containing 4,000 image sets provided by Shi et al.~\cite{shi2021towards} for our experiments where each image set consists of an RGB, depth map, amplitude, and phase image. We partitioned the 4,000 image sets into training, validation, and testing purposes of which amounts are 3,800, 100, and 100 respectively. Each model was trained for a total of 20 epochs with a fixed batch size of 4. The learning rate in the optimizer was set to 0.0001.

\subsection{Results}
\begin{figure}[ht]
  \centering
   \includegraphics[width=\linewidth]{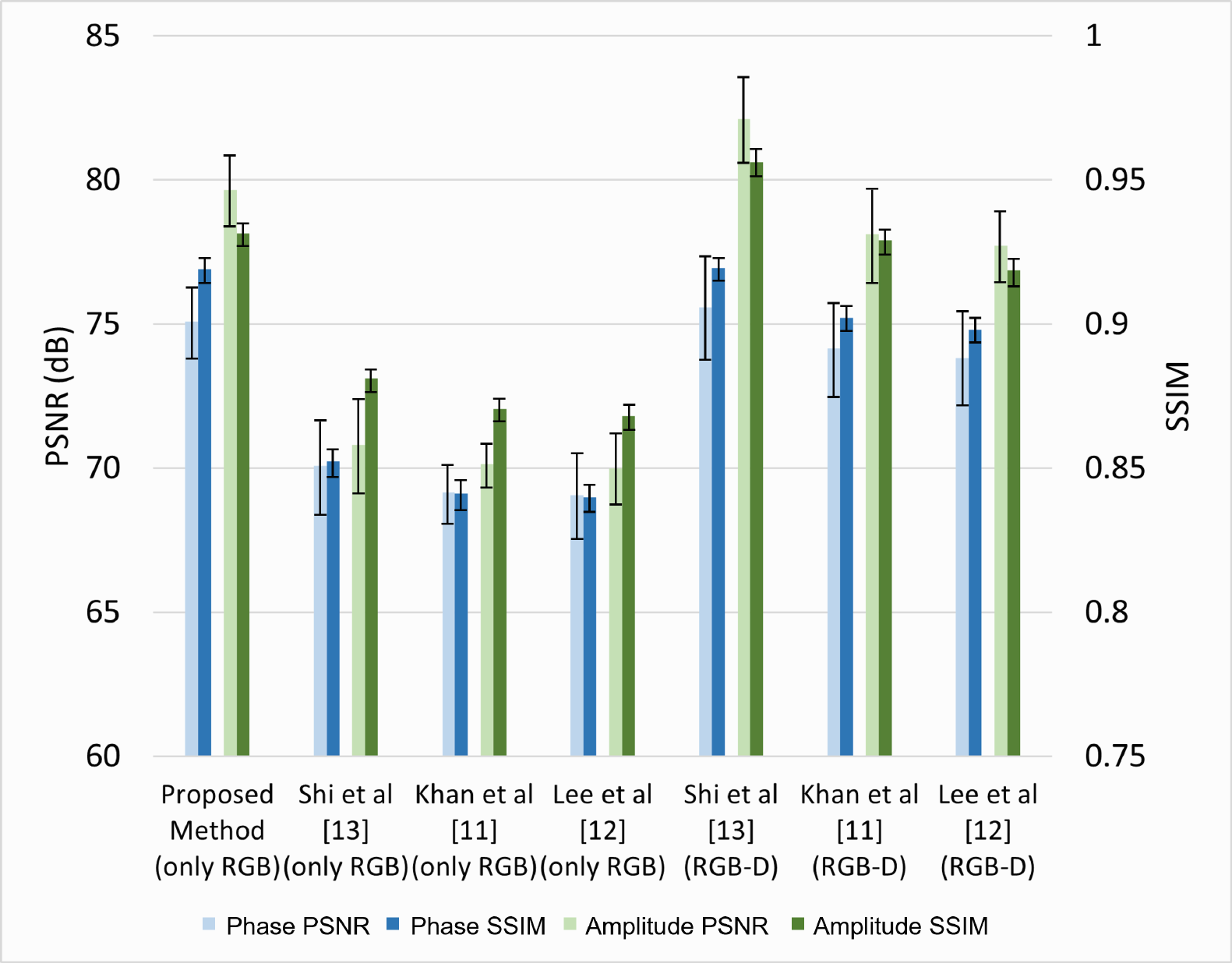}
   \caption{PSNR(dB), SSIM comparison of generated hologram images.}
   \label{fig5}
\end{figure}

\begin{figure}[ht]
  \centering
   \includegraphics[width=\linewidth]{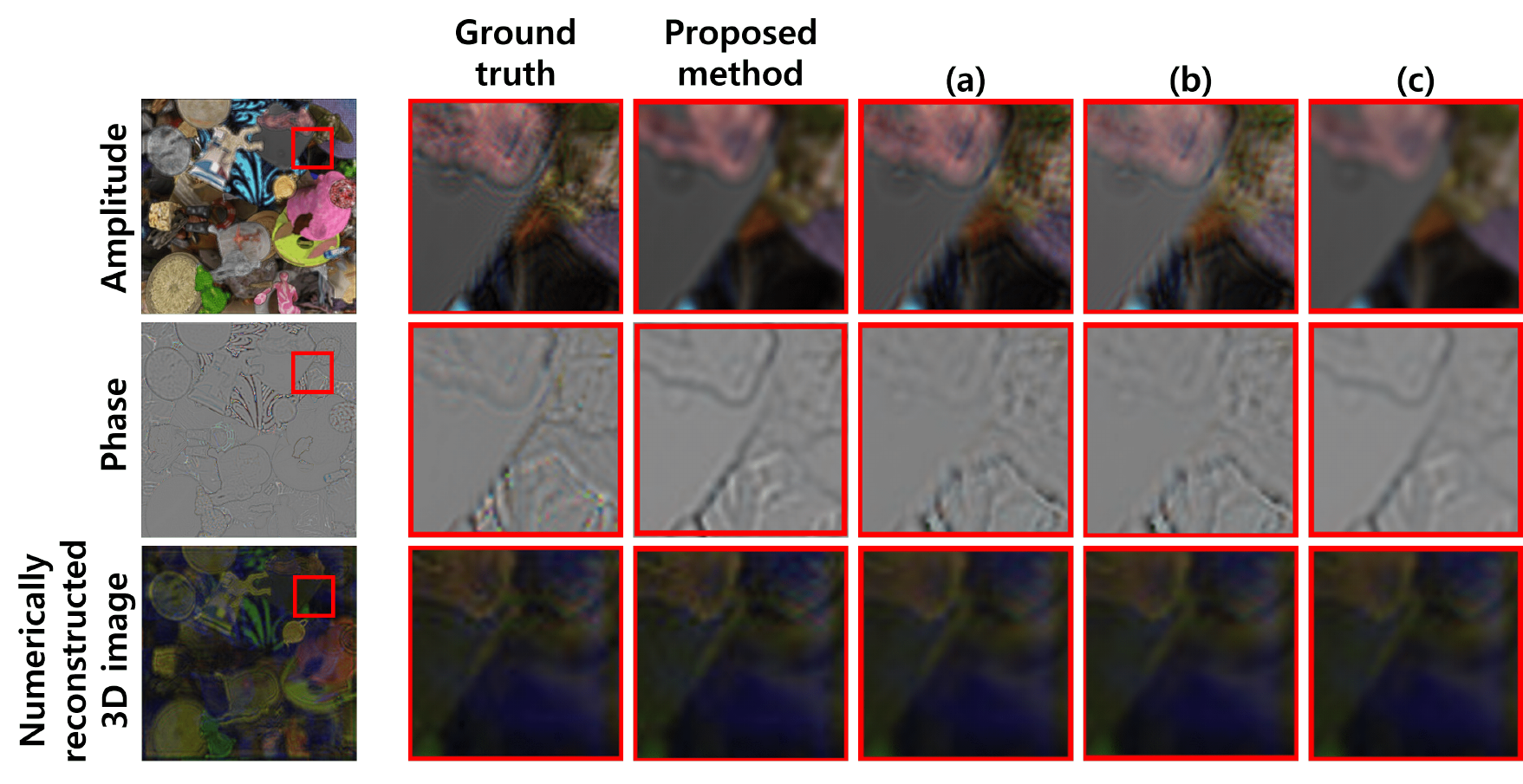}
   \caption{Qualitative differences between the proposed model and other models. (a): Shi et al \cite{shi2021towards}, (b): Lee et al \cite{lee2020deep}, (c): Khan et al \cite{khan2021gan}. In the numerically reconstructed 3D image,  camera's focus is on the enlarged part.}
   \label{fig6}
\end{figure}

Figure~\ref{fig5} shows the PSNR (Peak Signal-to-noise ratio) and SSIM of the four models. We found that the proposed model's results are more accurate than those of other models when every model used only an RGB image whereas the proposed model's results don't fall behind much than those of other models when other models fully use both RGB image and depth map. Table~\ref{tab:timecost} shows the measure of time required to train models when the proposed method uses only an RGB image whereas other models use fully use both an RGB image and depth map. The proposed method needs a significantly small amount of time than other models. This is due to the reduction of the number of parameters caused by the elimination of the depth map channel as input and the use of the Cascaded Convolutional blocks. Summarizing the numerical results, performances of the proposed method are slightly (about 1.57\%) lower than those of Shi et al\cite{shi2021towards}'s method, which is the best case in the experiment, however significantly (about 286\%) better in terms of time efficiency. Figure~\ref{fig6} shows visually qualitative differences, which are to use amplitude, phase, and reconstructed holographic 3D image, between the proposed model and other models when using only an RGB image. In the enlarged part including edge region information, it is visually confirmed that the edge part of the phase image from the proposed model is more similar to the ground truth than that of other models, and this behavior is maintained in the reconstructed holographic 3D image.

\begin{table}[ht]
\centering
\resizebox{\columnwidth}{!}{%
\begin{tabular}{c|c|c|c}
\hline
Proposed Method & Shi et al\cite{shi2021towards} & Khan et al\cite{khan2021gan} & Lee et al\cite{lee2020deep} \\ \hline
\textbf{2.9hr}           & 11.2hr    & 9.7hr      & 7.3hr     \\ \hline
\end{tabular}%
}
\caption{Measure of time to train models. The best result is in bold. (We measured the time from the training start to the end of 20 epochs using Nvidia's Titan RTX $\times$ 4.)}
\label{tab:timecost}
\end{table}

\section{Conclusion}
We proposed a deep learning method to generate the digital hologram without using depth map. The novel approach that consists of an Embedded Depth map Estimation module and CGH Generation module, using only RGB color information as input. Then we quantitatively and qualitatively compared the performance of the proposed method with the state-of-the-art models. Through our experiments, the following three main contributions are derived. First, PSNR and SSIM from the proposed model are more accurate than those from other competitive models when every model used only an RGB image. Also, the proposed model's PSNR and SSIM don't fall behind much than those of other models when other models fully use both RGB image and depth map. Second, the proposed method needs a significantly small amount of time for training than other competitive models. Third, we visually verify that the edge region of the phase image estimated from the proposed model gives more similarity with the ground truth than that of each competitive model, with this factor also maintained in the numerically reconstructed holographic 3D scenes. Considering the above demonstrations, the proposed method is expected to produce realistic holographic 3D content even under limited environments such as only RGB color information is available. It can also be a way to satisfy industrial needs that require large resolution and real-time hologram content such as MR/XR/metaverse platforms. We are planning a faster and more accurate CGH generation method by advancing depth map estimation and holographic 3D reconstruction.



\label{sec:refs}
\bibliographystyle{IEEEbib}
\bibliography{strings,refs}

\end{document}